\documentclass[a4paper,fleqn]{cas-dc}

\usepackage[numbers]{natbib}
\usepackage{microtype}

\usepackage{hyperref}
\hypersetup{colorlinks=true, citecolor=blue, linkcolor=blue, urlcolor=blue}
\usepackage[commandnameprefix=always]{changes}

\usepackage{changes}
\usepackage{todonotes}
\usepackage{xcolor}
\usepackage{float} 

\def\tsc#1{\csdef{#1}{\textsc{\lowercase{#1}}\xspace}}
\tsc{WGM}
\tsc{QE}

\usepackage{booktabs}       
\usepackage{nicefrac}       
\usepackage{microtype}      
\usepackage{lipsum}		
\usepackage[numbers]{natbib}
\usepackage{doi}
\usepackage{graphicx}   
\usepackage{caption}    
\usepackage{cleveref}   
\usepackage{array}
\usepackage{tabularx}
\usepackage{adjustbox}
\usepackage{enumitem}
\usepackage{longtable}
\usepackage{geometry}
\newcommand{\figref}[1]{\hyperref[#1]{Figure\ref*{#1}}}
\usepackage{longtable}
\usepackage{array}

\newcommand{\Algref}[1]{\hyperref[#1]{Algorithm\ref*{#1}}}
\begin{document}
\let\WriteBookmarks\relax
\def\floatpagepagefraction{1}
\def\textpagefraction{.001}
\let\printorcid\relax
\shorttitle{Optimizing Mortality Prediction for ICU Heart Failure Patients: Leveraging XGBoost and Advanced Machine Learning}   

\shortauthors{Negin Ashrafi et al.}

\title[mode = title]{Optimizing Mortality Prediction for ICU Heart Failure Patients: Leveraging XGBoost and Advanced Machine Learning with the MIMIC-III Database}   

\author[a]{Negin Ashrafi}
\author[b]{Armin Abdollahi}
\author[a]{Jiahong Zhang}
\author[a]{Maryam Pishgar}
\cormark[1]

\address[a]{Department of Industrial and Systems Engineering, University of Southern California (USC), 3650 McClintock Ave, Los Angeles, CA 90089, USA}
\address[b]{Department of Electrical and Computer Engineering, University of Southern California (USC), 3650 McClintock Ave, Los Angeles, CA 90089, USA}

\cortext[cor1]{Corresponding author.\\\hspace*{2em} \textit{E-mail address:} \href{mailto:pishgar@usc.edu}{pishgar@usc.edu} (Maryam Pishgar).}
\begin{abstract}
Heart failure affects millions of people worldwide, significantly reducing quality of life and leading to high mortality rates. Despite extensive research, the relationship between heart failure and mortality rates among ICU patients is not fully understood, indicating the need for more accurate prediction models. This study analyzed data from 1,177 patients over 18 years old from the MIMIC-III database, identified using ICD-9 codes. Preprocessing steps included handling missing data, removing duplicates, treating skewness, and using oversampling techniques to address data imbalances.

Through rigorous feature selection using Variance Inflation Factor (VIF), expert clinical input, and ablation studies, 46 key features were identified to enhance model performance. Our analysis compared several machine learning models, including Logistic Regression, Support Vector Machine (SVM), Random Forest, LightGBM, and XGBoost. XGBoost emerged as the superior model, achieving a test AUC-ROC of 0.9228 (95\% CI 0.8748 - 0.9613), significantly outperforming our previous work (AUC-ROC of 0.8766) and the best results reported in existing literature (AUC-ROC of 0.824).

The improved model's success is attributed to advanced feature selection methods, robust preprocessing techniques, and comprehensive hyperparameter optimization through Grid-Search. SHAP analysis and feature importance evaluations based on XGBoost highlighted key variables like leucocyte count and RDW, providing valuable insights into the clinical factors influencing mortality risk. This framework offers significant support for clinicians, enabling them to identify high-risk ICU heart failure patients and improve patient outcomes through timely and informed interventions.
\end{abstract}


\maketitle

\section{BACKGROUND}

Heart failure (HF) is a significant medical concern, affecting approximately 6.5 million Americans aged 20 years and older, making it one of the most prevalent cardiovascular conditions in the United States \cite{ref1}. As the inevitable outcome and final stage of many cardiac diseases, HF severely impacts the circulatory system, leading to symptoms such as shortness of breath, excessive coughing, and fatigue. These symptoms significantly affect patients' daily lives and are linked to an increased risk of early mortality, with about 25\% of HF cases resulting in death within one year \cite{ref1, ref2, ref3}. Approximately 20\% of hospitalized HF patients require ICU admission due to life-threatening complications, such as sepsis \cite{ref4, ref5}. Despite advanced care in ICU settings, in-hospital mortality rates for HF patients remain substantial, nearing 10\% \cite{ref6, ref7}. This persistent high mortality rate underscores the urgent need for robust predictive models to forecast HF patient outcomes in ICUs and enable timely medical interventions.

The implementation of Electronic Health Records (EHRs) has revolutionized the healthcare industry by providing comprehensive, digitally stored medical data, which enhances decision-making and increases the efficiency of patient care \cite{ref8, ref9, ref10, ref11}. EHRs allow for the systematic collection and analysis of patient data, facilitating data-driven decision-making processes that can optimize hospital performance \cite{ref12}. Machine Learning (ML) models have emerged as powerful tools for analyzing the vast amounts of data contained within EHRs, identifying complex patterns and correlations that are often undetectable using traditional statistical methods \cite{ref13, ref17, ref14, ref18}. These capabilities are particularly valuable in fields like cardiology, where the data is both extensive and highly varied \cite{ref16, ref19}.

Several prior studies have aimed to develop predictive models for mortality among HF patients, particularly those in ICUs \cite{ref20, ref21}. However, many of these models have struggled to achieve the level of reliability and accuracy necessary for clinical application. Effective feature selection and hyperparameter tuning are critical for enhancing model performance. Feature selection helps to identify the most relevant variables, reducing the risk of overfitting and improving model interpretability \cite{ref22}. Hyperparameter tuning involves adjusting the model's preset parameters to optimize performance for specific clinical scenarios, enhancing both the precision and robustness of predictions \cite{ref23}. By combining these methods, it is possible to create models that are both computationally efficient and highly accurate, making them valuable tools for clinical decision support.

This study aims to improve the prediction of in-hospital mortality for ICU patients with HF by implementing innovative preprocessing and feature selection techniques. We employed systematic imputation strategies, using either the mean or median depending on the distribution of each variable, and conducted univariate analyses using VIF and XGBoost feature selection method. We used advanced machine learning models, such as XGBoost, to predict ICU mortality among heart failure patients. XGBoost, a robust gradient boosting algorithm, is highly effective in handling structured data and capturing complex patterns. By iteratively correcting errors, it outperforms traditional models like Logistic Regression and Decision Trees \cite{ref28, ref29}. These methodologies significantly enhanced the Area Under the Curve-Receiver Operating Characteristic (AUC-ROC) of our models, achieving results that surpass the benchmarks established in previous studies \cite{ref25, ref27}. Our research adheres to the Transparent Reporting of a Multivariable Prediction Model for Individual Prognosis or Diagnosis (TRIPOD) guidelines, ensuring the transparency and reproducibility of our findings \cite{ref26, ref24}. This paper extends our previous work by including additional explanations, improved experimental results, and more comprehensive analysis \cite{ref15}.

\section{METHODOLOGY}

\subsection{\textit{Data Source and Study Design}}
The MIMIC-III (version 1.4) database is an extensive, publicly accessible database that recorded 38,597 adult patients and 49,785 hospital admissions who stayed in ICUs of the Beth Israel Deaconess Medical Center in Boston, Massachusetts from 2001 to 2012 \cite{ref23}. This dataset includes information on admissions, patient demographics, vital sign measurements, laboratory test results, procedures, medications, caregiver notes, imaging reports, and mortality (including dates and times). This comprehensive dataset supports extensive research in clinical informatics. We chose MIMIC-III for its substantial real-world data, which enhances our research's depth. After data extraction, preprocessing ensured quality and suitability for model training, providing a robust foundation for predictive modeling and clinical insights.

\subsection{\textit{Patient extraction}}

\begin{figure}[!htb]
    \centering
    \includegraphics[width=\linewidth]{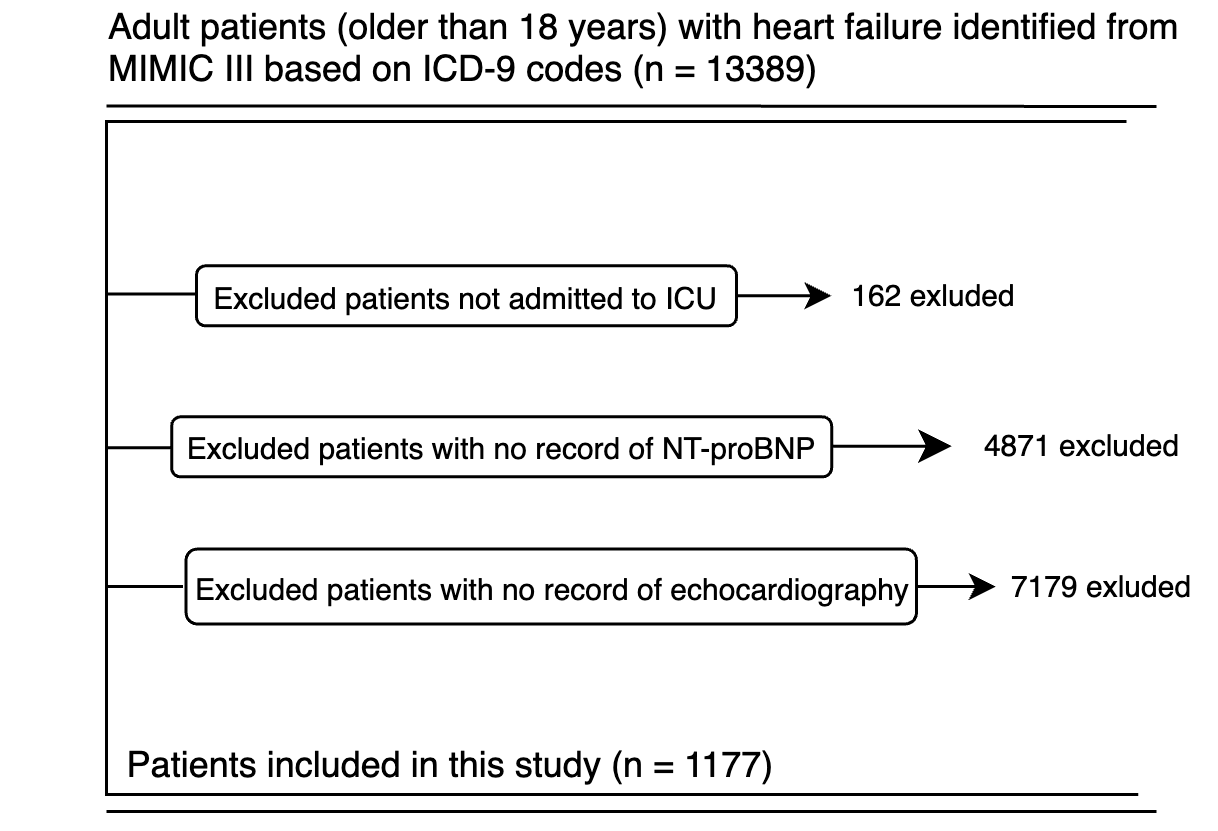}
    \caption{Patient selection process for heart failure study, showing exclusions and final cohort size.}
    \label{fig:1}
\end{figure}

Our study focused on extracting data for adult patients diagnosed with heart failure from the MIMIC-III database, specifically targeting individuals over 18 identified by relevant ICD-9 codes. This initial selection comprised 13,389 patients. To refine the dataset for targeted analysis, we applied several exclusion criteria: we removed 162 patients without ICU admissions, as our focus was on ICU care. Additionally, 4,871 patients were excluded due to missing NT-proBNP records, a key heart failure biomarker \cite{ref30}, and 7,179 were excluded for lacking echocardiography records, critical for heart failure diagnosis \cite{ref31}. These steps reduced our final cohort to 1,177 patients, providing a focused dataset to gain specific insights into heart failure management in critical care. We then divided this cohort into training and test sets, ensuring robust model training and unbiased performance evaluation. \autoref{fig:1} illustrates the data extraction process, showing the progression from initial selection to the final cohort.

\subsection{\textit{Feature selection}}
Using Structured Query Language (SQL) with PostgreSQL (V.9.6), a comprehensive set of demographic characteristics, vital signs, and laboratory values were initially extracted from the MIMIC-III dataset. This large set of features was based on insights from previous studies (\cite{ref2, ref32, ref33}), clinical relevance, and expert opinions. Demographic characteristics and vital signs were recorded during the first 24 hours of each admission, while laboratory variables were measured throughout the entire ICU stay. Mean values were analyzed for features with multiple measurements. Here, \autoref{tab:1} shows the list of features categorized into demographic characteristics, vital signs, comorbidities, and laboratory variables.

Given the extensive number of features initially considered, a more focused approach was necessary to ensure model effectiveness and clinical relevance. To prevent multicollinearity among continuous features, the Variance Inflation Factor (VIF) was calculated, and variables with a VIF exceeding the threshold of 5 were removed, reducing potential issues with high standard errors in the prediction model \cite{ref34}.  

\begin{figure}[!htb]
    \centering
    \includegraphics[width=\linewidth]{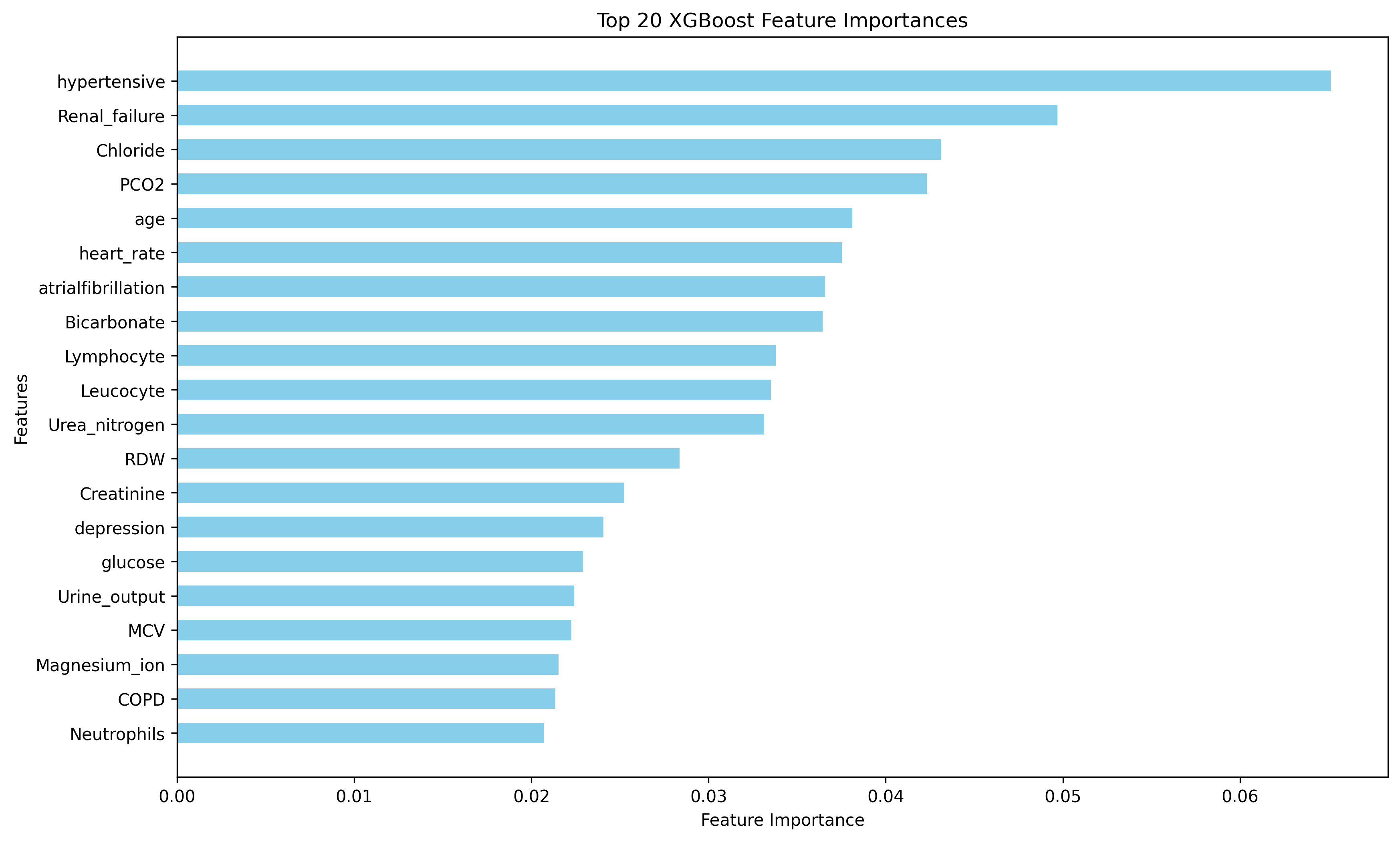}
    \caption{Top 20 feature importances ranked by XGBoost, showing 'hypertensive' and 'Renal\_failure' as the most influential features.}
    \label{fig:2}
\end{figure}

Through expert recommendations and further analysis, the extensive set of features was refined to focus on those most relevant to the model's performance and clinical outcomes. This process resulted in a selection of 48 features for our models, providing a balanced and effective set of predictors for in-hospital mortality. To further understand the contribution of each feature, XGBoost feature importance was utilized to extract the top 20 contributing features, highlighting the most impactful variables based on this model. Here, \autoref{fig:2} shows the feature importance ranking by XGBoost.

\begin{table*}[width = .8\textwidth,pos=!htb]
    \centering
    \caption{Categories and variables used in the study.}
    \begin{tabular*}{\tblwidth}{@{\extracolsep{\fill}}l p{10cm}}
    \midrule
         \textbf{Category} & \textbf{Variable} \\
    \midrule
        \textbf{Demographic Characteristics} & Age, Gender, BMI (Body Mass Index) \\
    \midrule
        \textbf{Vital Signs} & Heart rate (HR), Systolic blood pressure (SBP), \\
                              & Diastolic blood pressure (DBP), Mean blood pressure, \\
                              & Respiratory rate, Temperature, SPO2 (Saturation pulse oxygen), \\
                              & Urine output (first 24 hours) \\
    \midrule
        \textbf{Comorbidities} & Hypertension, Atrial fibrillation, \\
                               & CHD with no MI (Coronary Heart Disease without Myocardial Infarction), \\
                               & Diabetes mellitus, Depression, Deficiencies anemias, \\
                               & Hyperlipemia (Hyperlipidaemia), Chronic kidney disease (CKD) (Renal failure), \\
                               & Chronic obstructive pulmonary disease (COPD) \\
    \midrule
        \textbf{Laboratory Variables} & Hematocrit, Red Blood Cells (RBC), \\
                                      & Mean Corpuscular Hemoglobin (MCH), \\
                                      & Mean Corpuscular Hemoglobin Concentration (MCHC), \\
                                      & Mean Corpuscular Volume (MCV), \\
                                      & Red Cell Distribution Width (RDW), \\
                                      & White Blood Cells (Leucocyte), Platelet count, Neutrophils, \\
                                      & Basophils, Lymphocytes, Prothrombin Time (PT), \\
                                      & International Normalized Ratio (INR), \\
                                      & N-terminal pro b-type Natriuretic Peptide (NT-proBNP), \\
                                      & Creatine Kinase, Creatinine, Blood Urea Nitrogen (BUN), \\
                                      & Glucose, Blood Potassium, Blood Sodium, Blood Calcium, \\
                                      & Chloride, Anion gap, Magnesium, \\
                                      & Hydrogen ion concentration (pH), \\
                                      & Bicarbonate, Lactic acid, Partial pressure of CO2 (PCO2) \\
    \midrule
    \end{tabular*}
    \label{tab:1}
\end{table*}

\subsection{\textit{Data Preprocessing}}

Our study began with preprocessing the raw data extracted from the MIMIC-III database, focusing on adult patients diagnosed with heart failure. Initial steps involved reading and cleaning the dataset by excluding irrelevant variables such as 'group' and 'ID,' removing duplicate entries, and eliminating columns with only a single unique value. Rows with missing outcome data were discarded due to the absence of real-world outcomes, ensuring the integrity of the dataset for analysis.

To address missing values, we utilized median imputation, chosen for its robustness against the skewness and outliers present in most features. This approach ensured that the imputed values did not disproportionately affect the dataset's distribution. Outliers were further managed by selectively removing extreme data points, which helped maintain the accuracy and reliability of our findings.

\begin{figure*}[!htb]
    \centering
    \includegraphics[width=15cm, height=10cm]{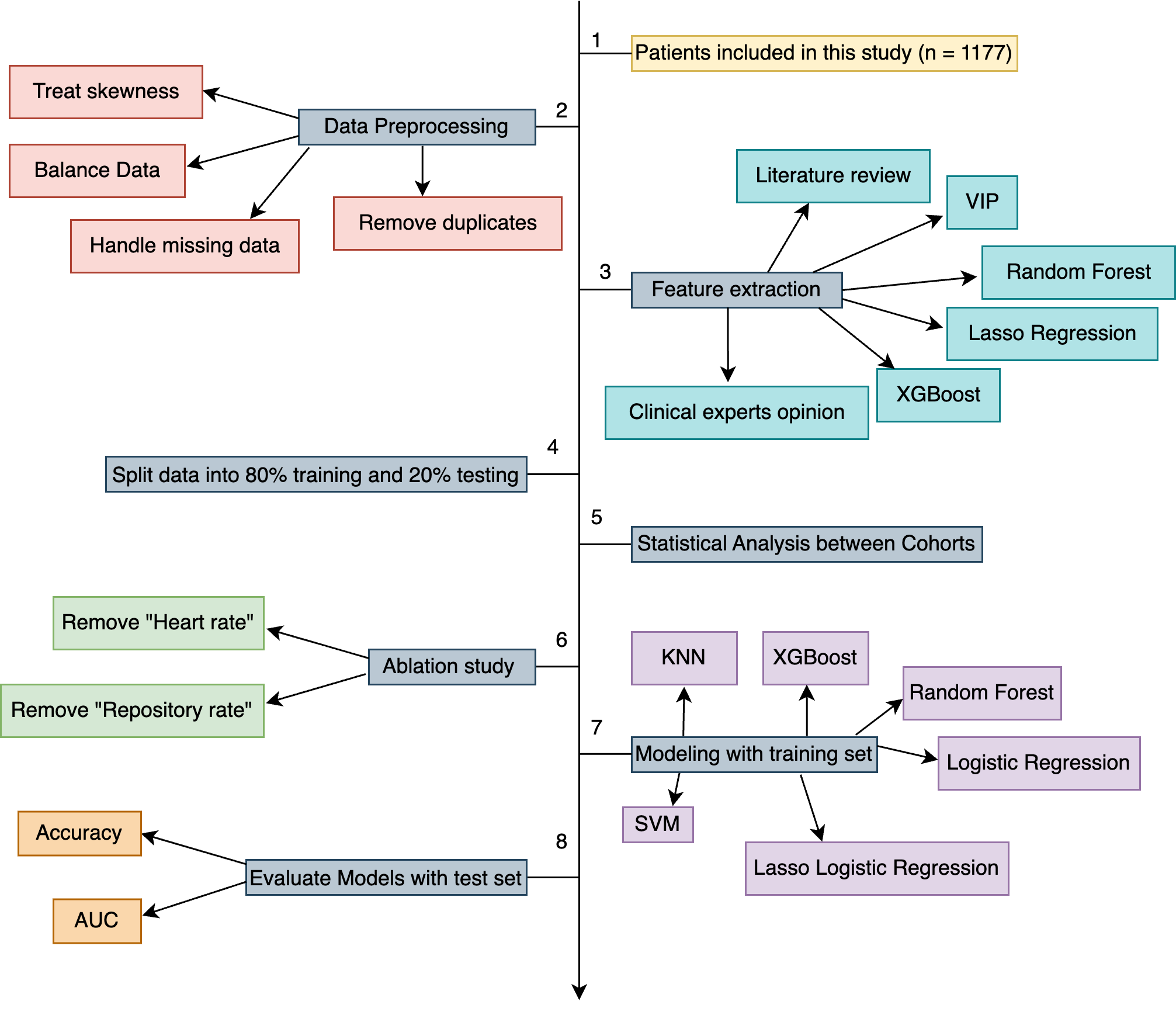}
    \caption{Overview of the study workflow, including data preprocessing, feature extraction, modeling, and evaluation.}
    \label{fig:3}
\end{figure*}

After handling missing values and outliers, we assessed the distribution of classes within the outcome variable, revealing an imbalance. To rectify this, we implemented oversampling techniques, particularly in the training set, to balance the classes and improve model training efficacy. Following these steps, the dataset was divided into training and test sets, comprising 80\% and 20\% of the data, respectively. This thorough preprocessing laid a solid foundation for subsequent feature selection and predictive modeling. The whole process is summarized in \autoref{fig:3}.

\subsection{\textit{Modeling}}

We utilized a suite of machine learning models to predict ICU mortality in heart failure patients: Logistic Regression, LASSO Logistic Regression, Random Forest, LightGBM, Support Vector Machine, and XGBoost. These models were selected for their proven effectiveness in mortality prediction, as supported by existing literature. Many were used in our prior work, and XGBoost was added to enhance performance. XGBoost is particularly powerful in this field due to its ability to handle complex data and improve predictive accuracy through gradient boosting.

Hyperparameter optimization was conducted using Grid-Search to find the optimal settings for each model. Model performance was assessed by calculating AUC-ROC values and accuracy on the test set. Evaluation criteria included bootstrapped 95\% confidence intervals for both AUC-ROC and accuracy, with higher AUC-ROC values indicating better discrimination. XGBoost achieved the highest AUC-ROC and accuracy, demonstrating superior predictive power and robustness in identifying ICU mortality risk among heart failure patients. This modeling process provided robust insights into the data, facilitating effective prediction of patient outcomes.

\section{RESULTS}

\subsection{\textit{Statistical Analysis between Cohorts}}

A thorough statistical analysis was performed to evaluate the comparability between our training and testing datasets. Using an independent two-sample t-test, we assessed whether there were significant differences in the means of various features between the two groups \cite{ref35}. As detailed in \autoref{tab:2}, out of the 34 features examined, 31 had p-values above the 0.05 significance threshold, indicating no substantial differences between the training and testing sets for these features. This suggests that the training and testing cohorts are well-matched, supporting the reliability of our model’s predictive capabilities. However, a few features, such as 'RBC,' 'MCV,' and 'Blood\_calcium,' exhibited significant differences, with p-values below 0.05, indicating potential variations in patient characteristics or data collection methods. Despite these differences, the overall consistency between the cohorts underscores the robustness of our model, ensuring its ability to generalize effectively across diverse data subsets.

\begin{table*}[width = .8\textwidth,pos=!htb]
    \centering
    \caption{Comparison of Train and Test Means (Std) with p-values across various features.}
    \begin{tabular*}{\tblwidth}{@{\extracolsep{\fill}}lccc}
    \midrule
         \textbf{Feature} & \textbf{Train Mean (Std)} & \textbf{Test Mean (Std)} & \textbf{p-value} \\
    \midrule
        Age (years) & 75.57 (12.13) & 75.59 (11.83) & 0.9832 \\
        BMI & 28.66 (6.00) & 29.39 (6.44) & 0.1442 \\
        Heart rate (HR) & 83.43 (15.45) & 83.78 (15.72) & 0.7734 \\
        Systolic blood pressure (mmHg) & 117.34 (16.72) & 117.92 (14.55) & 0.6241 \\
        Diastolic blood pressure (mmHg) & 58.74 (10.03) & 58.63 (9.26) & 0.8871 \\
        Respiratory rate (breaths/min) & 20.49 (3.78) & 21.00 (3.53) & 0.0705 \\
        Temperature (°C) & 36.66 (0.56) & 36.70 (0.60) & 0.3881 \\
        SPO2 (\%) & 96.46 (2.03) & 96.19 (2.05) & 0.0968 \\
        Urine output (ml/24h) & 1807.35 (1063.12) & 1824.25 (1067.17) & 0.8404 \\
        Hematocrit (\%) & 31.63 (4.78) & 32.28 (5.27) & 0.1087 \\
        RBC (million cells/\textmu L) & 3.52 (0.57) & 3.63 (0.58) & 0.0170 \\
        MCH (pg/cell) & 29.77 (2.41) & 29.41 (2.54) & 0.0623 \\
        MCHC (g/dL) & 32.94 (1.29) & 32.92 (1.42) & 0.8557 \\
        MCV (fL) & 90.44 (6.13) & 89.35 (6.12) & 0.0246 \\
        RDW (\%) & 15.88 (1.97) & 15.64 (1.97) & 0.1292 \\
        Leucocyte (cells/\textmu L) & 10.51 (4.30) & 10.42 (4.14) & 0.7959 \\
        Platelets (cells/\textmu L) & 239.89 (107.09) & 246.64 (96.68) & 0.3861 \\
        Neutrophils (\%) & 81.29 (8.60) & 81.13 (8.88) & 0.8159 \\
        Basophils (\%) & 0.36 (0.24) & 0.35 (0.24) & 0.6412 \\
        Lymphocyte (\%) & 11.46 (6.70) & 11.62 (7.29) & 0.7816 \\
        PT (seconds) & 17.25 (6.63) & 17.19 (6.83) & 0.9102 \\
        INR & 1.60 (0.75) & 1.60 (0.78) & 0.9368 \\
        NT-proBNP (pg/mL) & 11156.64 (12705.77) & 10818.28 (13490.49) & 0.7472 \\
        Creatine kinase (U/L) & 164.46 (321.86) & 197.62 (343.73) & 0.2152 \\
        Creatinine (mg/dL) & 1.59 (1.06) & 1.59 (1.07) & 0.9604 \\
        Urea nitrogen (mg/dL) & 36.26 (20.92) & 35.00 (21.00) & 0.4488 \\
        Glucose (mg/dL) & 147.29 (48.92) & 147.88 (47.41) & 0.8752 \\
        Blood Potassium (mmol/L) & 4.16 (0.40) & 4.13 (0.35) & 0.3911 \\
        Blood Sodium (mmol/L) & 139.09 (3.88) & 139.20 (3.69) & 0.7084 \\
        Blood Calcium (mg/dL) & 8.49 (0.56) & 8.59 (0.56) & 0.0183 \\
        Chloride (mmol/L) & 102.72 (5.07) & 102.38 (5.07) & 0.3849 \\
        Anion gap (mmol/L) & 13.75 (2.37) & 13.87 (2.40) & 0.5454 \\
        Magnesium ion (mg/dL) & 2.12 (0.25) & 2.10 (0.22) & 0.2455 \\
        PH & 7.38 (0.06) & 7.38 (0.06) & 0.6166 \\
        Bicarbonate (mmol/L) & 26.83 (4.83) & 27.16 (4.86) & 0.3991 \\
        Lactic acid (mmol/L) & 1.75 (0.73) & 1.68 (0.67) & 0.2167 \\
        PCO2 (mmHg) & 44.53 (10.22) & 44.73 (10.32) & 0.8091 \\
        EF (\%) & 48.78 (12.87) & 49.53 (12.45) & 0.4437 \\
    \midrule
    \end{tabular*}
    \label{tab:2}
\end{table*}

\subsection{\textit{Ablation Study}}

\begin{figure*}[!htb]
    \centering
    \includegraphics[width=15cm, height=7.5cm]{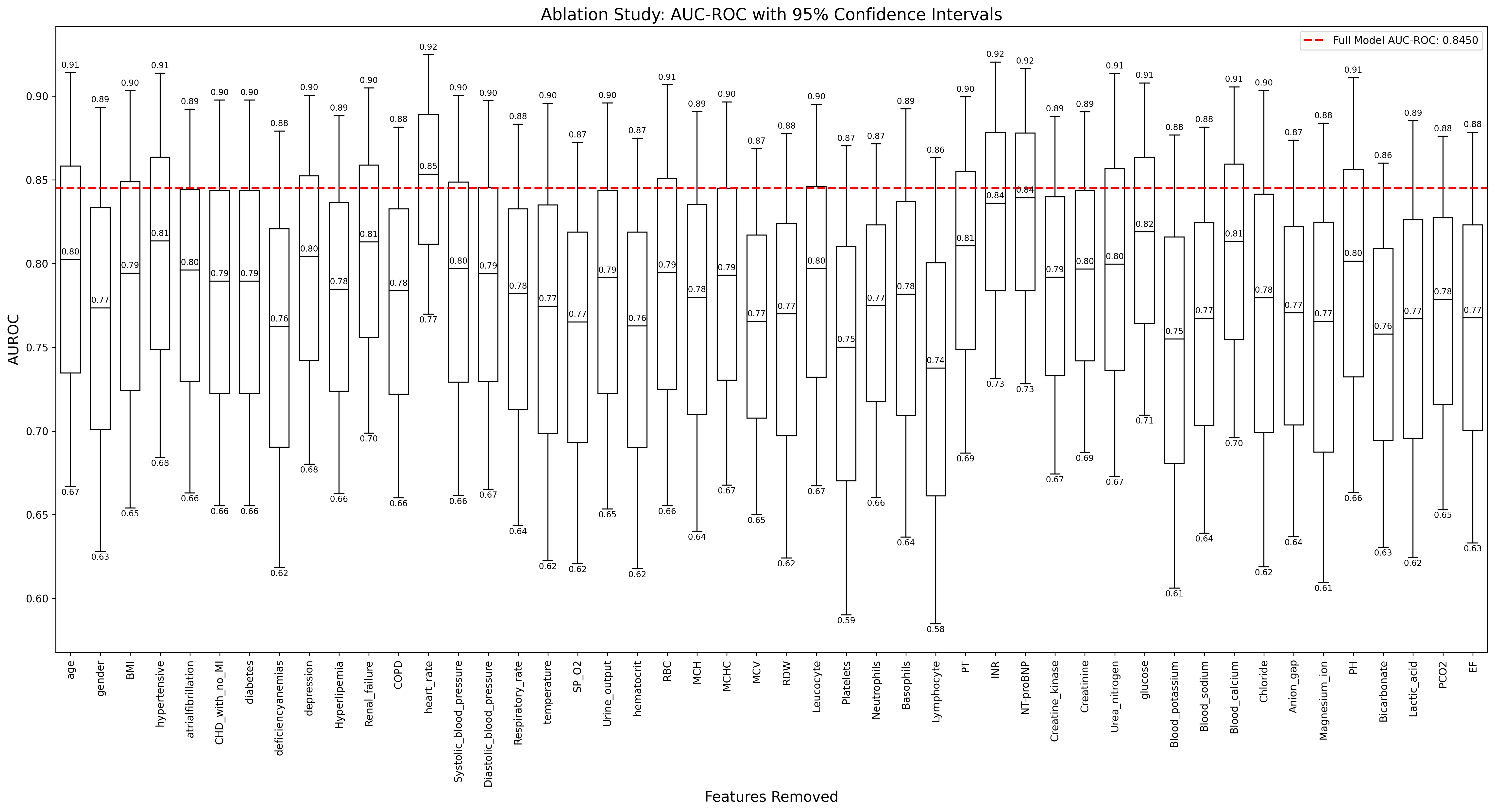}
    \caption{AUC boxplot of the XGBoost model on the test dataset with all features included, achieving an AUC of 0.8450.}
    \label{fig:4}
\end{figure*}

\begin{figure*}[!htb]
    \centering
    \includegraphics[width=15cm, height=7.5cm]{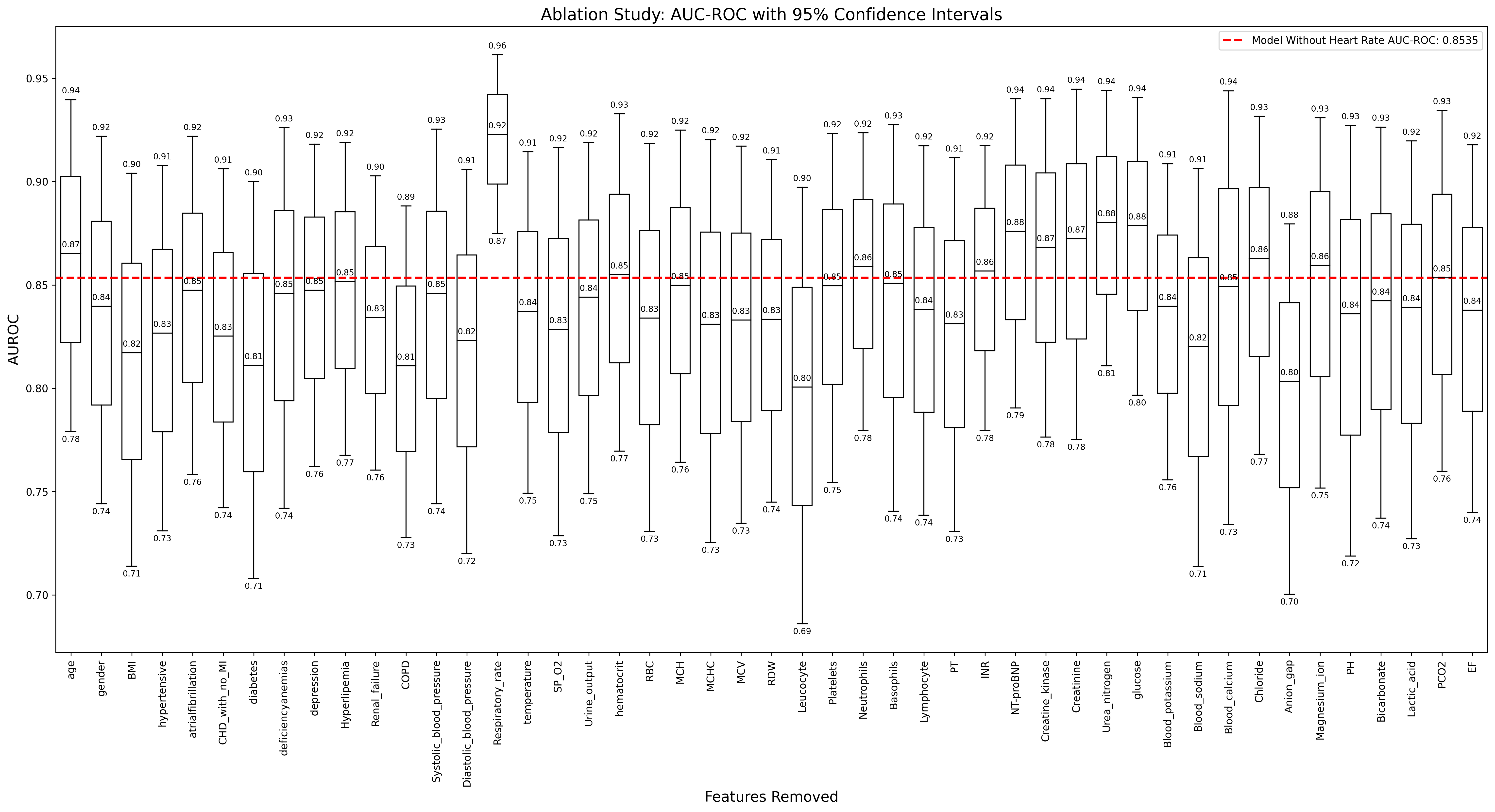}
    \caption{AUC boxplot of the XGBoost model on the test dataset after removing the 'Heart Rate' feature, showing an increase in AUC to 0.8535.}
    \label{fig:5}
\end{figure*}

The ablation study was conducted to evaluate the impact of individual physiological features on the model's predictive performance, as measured by the Area Under the Curve (AUC). The baseline model, incorporating a comprehensive set of features, initially achieved an AUC of approximately 0.8450, as illustrated in \autoref{fig:4}. Interestingly, the removal of the 'Heart Rate' feature led to a slight increase in AUC to 0.8535, suggesting that 'Heart Rate' might not be a critical predictor of the target outcome in this specific clinical context. This result, depicted in \autoref{fig:5}, implies that while 'Heart Rate' is a standard clinical measure, its contribution to the predictive model may be minimal, potentially due to its variability or interaction with other more informative features.

\begin{figure*}[!htb]
    \centering
    \includegraphics[width=15cm, height=7.5cm]{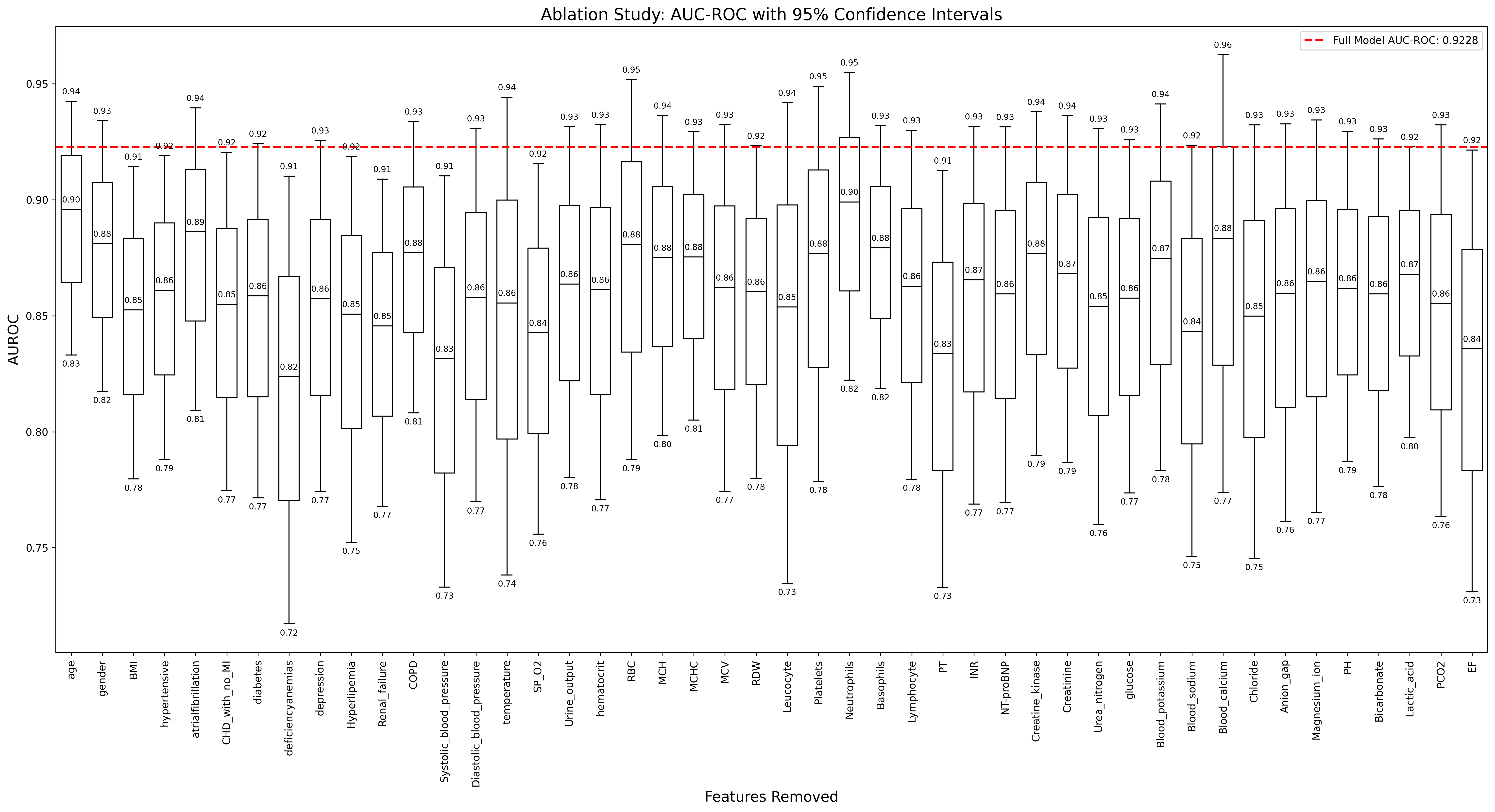}
    \caption{AUC boxplot of the XGBoost model on the test dataset after removing both 'Heart Rate' and 'Respiratory Rate' features, demonstrating an improved AUC of 0.9228.}
    \label{fig:6}
\end{figure*}

Further refinement of the model involved the removal of the 'Respiratory Rate' feature, in addition to 'Heart Rate.' This adjustment led to a substantial improvement in model performance, raising the AUC to 0.9228, as shown in \autoref{fig:6}. This significant increase suggests that both 'Heart Rate' and 'Respiratory Rate' may introduce noise or exhibit high collinearity with other features, thereby reducing their individual predictive value. The exclusion of these features optimized the model by minimizing redundancy and enhancing the signal clarity derived from the remaining features, which are more directly associated with the patient's clinical outcomes.

By identifying and removing the less impactful features, specifically 'Heart Rate' and 'Respiratory Rate', the optimized feature set achieved the highest observed AUC, confirming that no further feature removal was required. This ablation study underscores the importance of rigorous feature selection in clinical predictive modeling, where careful evaluation of physiological variables can lead to significant improvements in model robustness and accuracy. These findings highlight the need for continuous assessment of commonly used clinical indicators and their relevance in various predictive contexts, ensuring that models are both efficient and effective in identifying critical patient outcomes.

\subsection{\textit{Evaluation results}}

\autoref{tab:3} and \autoref{tab:4} summarize the results of our proposed model and baseline ML models using our evaluation metrics for both the training and test datasets. XGBoost emerged as the top-performing model, achieving an accuracy of 0.9993 and an AUC-ROC of 1.0000 (95\% CI 1.0000 - 1.0000) on the training set, and an accuracy of 0.8966 with an AUC-ROC of 0.9228 (95\% CI 0.8748 - 0.9613) on the test set, demonstrating its robust capability in predicting ICU mortality in heart failure patients.

\begin{figure*}[!htb]
    \centering
    \includegraphics[width=13cm, height=9cm]{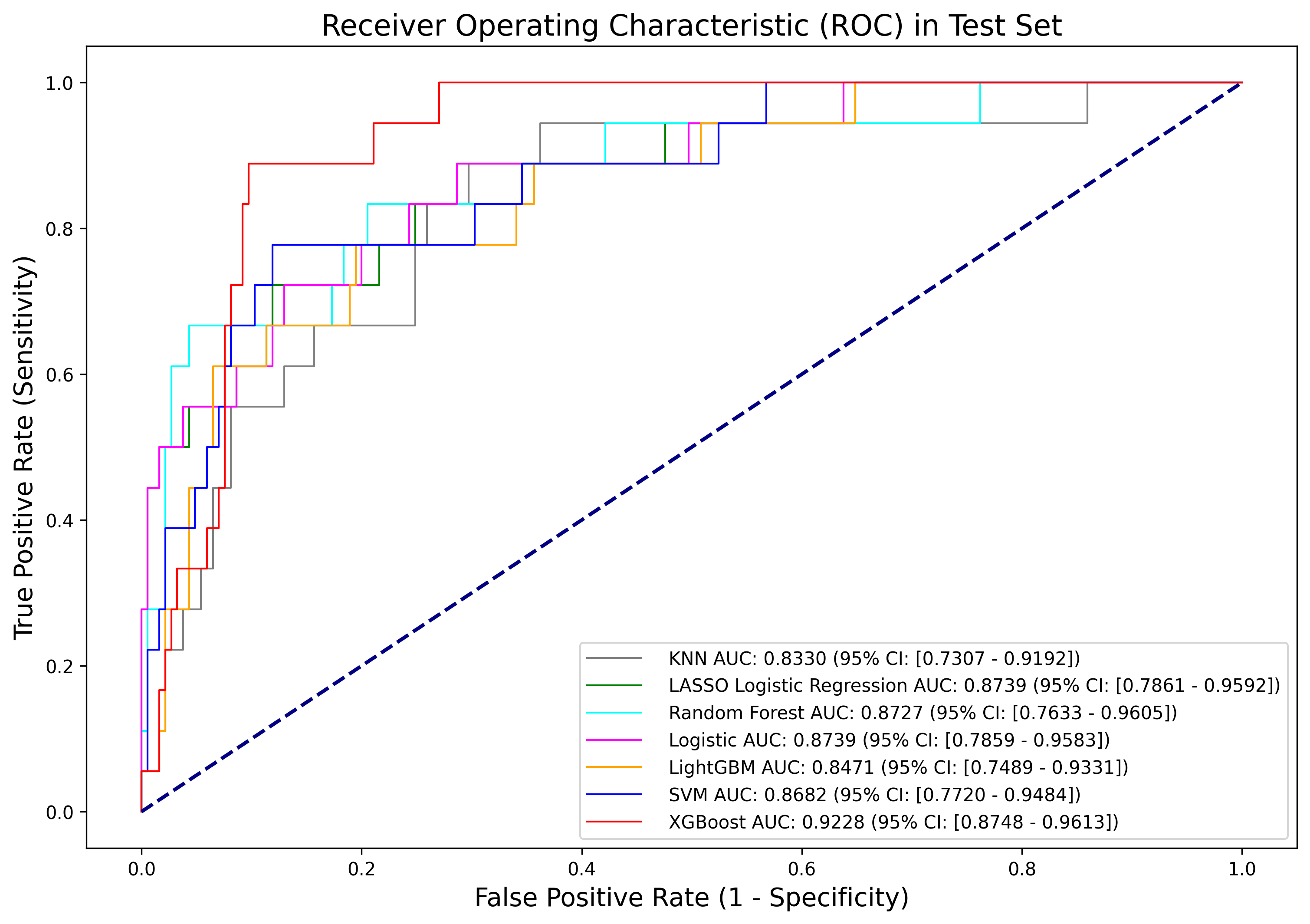}
    \caption{ROC curves for various models tested on the test dataset. The XGBoost model achieved the highest AUC of 0.9228, indicating superior predictive performance compared to other models.}
    \label{fig:7}
\end{figure*}

Following XGBoost, the Random Forest and LightGBM models ranked second and third, respectively. Random Forest achieved an accuracy of 1.0000 and an AUC-ROC of 1.0000 (95\% CI 1.0000 - 1.0000) on the training set, and an accuracy of 0.9212 with an AUC-ROC of 0.8727 (95\% CI 0.7633 - 0.9605) on the test set. LightGBM also showed strong performance, with an accuracy of 1.0000 and an AUC-ROC of 1.0000 (95\% CI 1.0000 - 1.0000) on the training set, and an accuracy of 0.9015 with an AUC-ROC of 0.8471 (95\% CI 0.7489 - 0.9331) on the test set. These results highlight that, while traditional models provide reasonable accuracy, advanced models like XGBoost, Random Forest, and LightGBM significantly enhance predictive capabilities in complex healthcare datasets. \autoref{fig:7} illustrates the AUC curves for the test set, highlighting the discriminative ability of these models in identifying ICU mortality risks.

\begin{table*}[width = .8\textwidth,pos=!htb]
    \centering
    \caption{Comparison of Models Based on AUC, Confidence Interval, and Accuracy (Train Dataset)}
    \begin{tabular*}{\tblwidth}{@{\extracolsep{\fill}}lccc}
    \midrule
       Models	& AUROC	& AUROC 95\% CI	& Accuracy\\
       \midrule
XGBoost & 1.0000 & [1.0000 - 1.0000] & 0.9993 \\
Logistic Regression & 0.8395 & [0.8216 - 0.8586] & 0.7559 \\
Random Forest & 1.0000 & [1.0000 - 1.0000] & 1.0000 \\
LightGBM & 1.0000 & [1.0000 - 1.0000] & 1.0000 \\
SVM & 0.9994 & [0.9989 - 0.9998] & 0.9845 \\
Lasso & 0.8389 & [0.8206 - 0.8580] & 0.7573 \\
KNN & 1.0000 & [1.0000 - 1.0000] & 1.0000 \\
\midrule
    \end{tabular*}
    \label{tab:3}
\end{table*}

\begin{table*}[width = .8\textwidth,pos=!htb]
    \centering
    \caption{Comparison of Models Based on AUC, Confidence Interval, and Accuracy (Test Dataset)}
    \begin{tabular*}{\tblwidth}{@{\extracolsep{\fill}}lccc}
    \midrule
       Models	& AUROC	& AUROC 95\% CI	& Accuracy\\
       \midrule
XGBoost & 0.9228 & [0.8748 - 0.9613] & 0.8966 \\
Logistic Regression & 0.8739 & [0.7859 - 0.9583] & 0.8079 \\
Random Forest & 0.8727 & [0.7633 - 0.9605] & 0.9212 \\
LightGBM & 0.8471 & [0.7489 - 0.9331] & 0.9015 \\
SVM & 0.8682 & [0.7720 - 0.9484] & 0.8966 \\
Lasso & 0.8739 & [0.7861 - 0.9592] & 0.8079 \\
KNN & 0.8330 & [0.7307 - 0.9192] & 0.8818 \\
\midrule
    \end{tabular*}
    \label{tab:4}
\end{table*}

\subsection{\textit{SHAP analysis}}

\begin{figure*}[!htb]
    \centering
    \includegraphics[width=11cm, height=8cm]{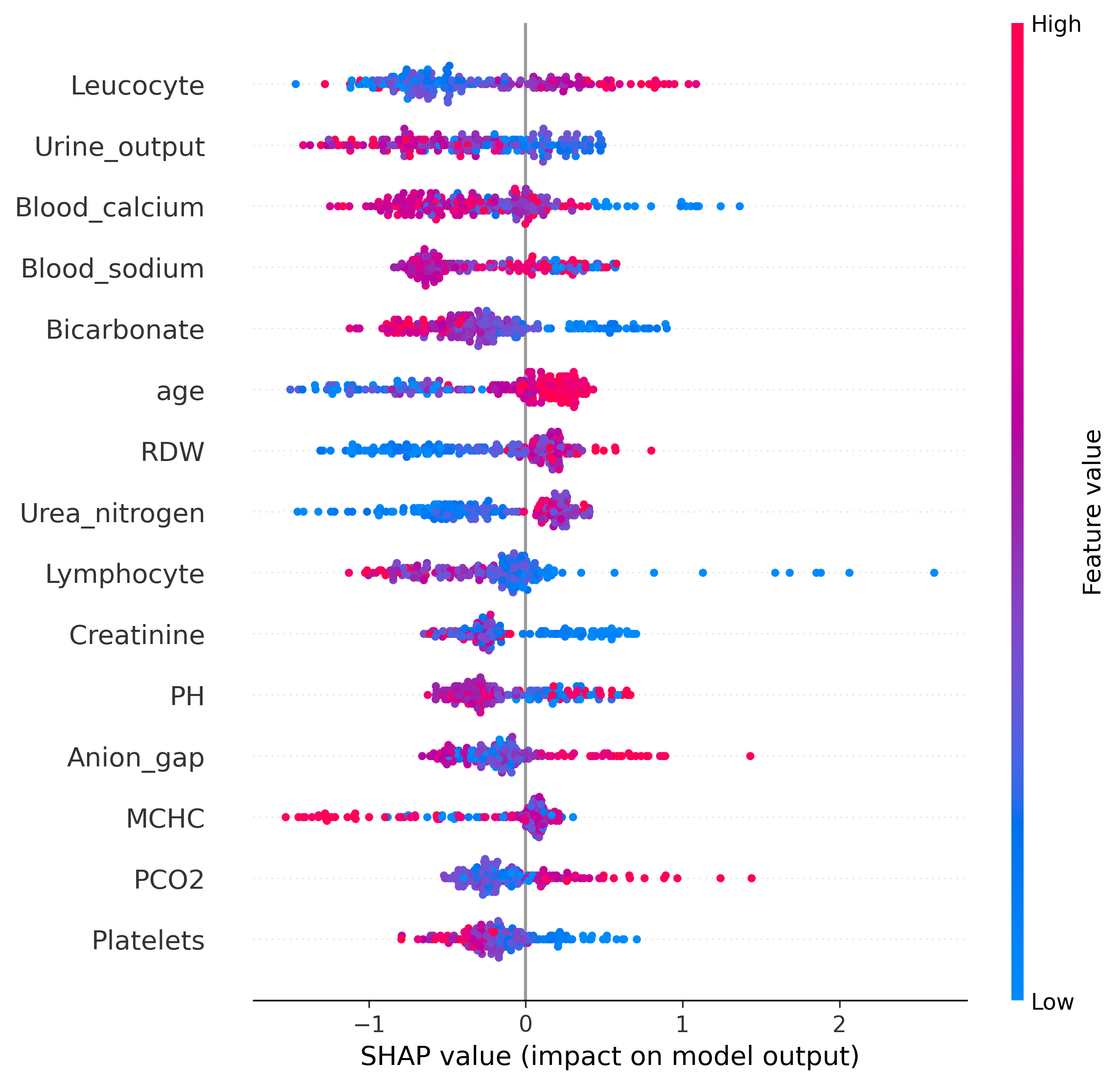}
    \caption{SHAP summary plot showing the impact of the top features on the XGBoost model's output, with colors representing feature values (red for high, blue for low).}
    \label{fig:8}
\end{figure*}

SHAP analysis was utilized to identify the most influential features impacting the ML model's predictions, as suggested by Hamilton et al. \cite{ref36}. The SHAP summary plot (Figure 8) and the mean SHAP values (\autoref{fig:8}) illustrate the top features in descending order of importance. 'Leucocyte' emerged as the most critical feature for predicting ICU mortality in heart failure patients. Features such as 'Leucocyte,' 'Urine\_output,' and 'Blood\_calcium' showed high SHAP values, indicating a significant impact on the model's output. Higher SHAP values for 'Leucocyte' and 'RDW' corresponded to an increased risk of mortality, while higher 'Urine\_output' values were associated with a lower risk.

\begin{figure*}[!htb]
    \centering
    \includegraphics[width=11cm, height=8cm]{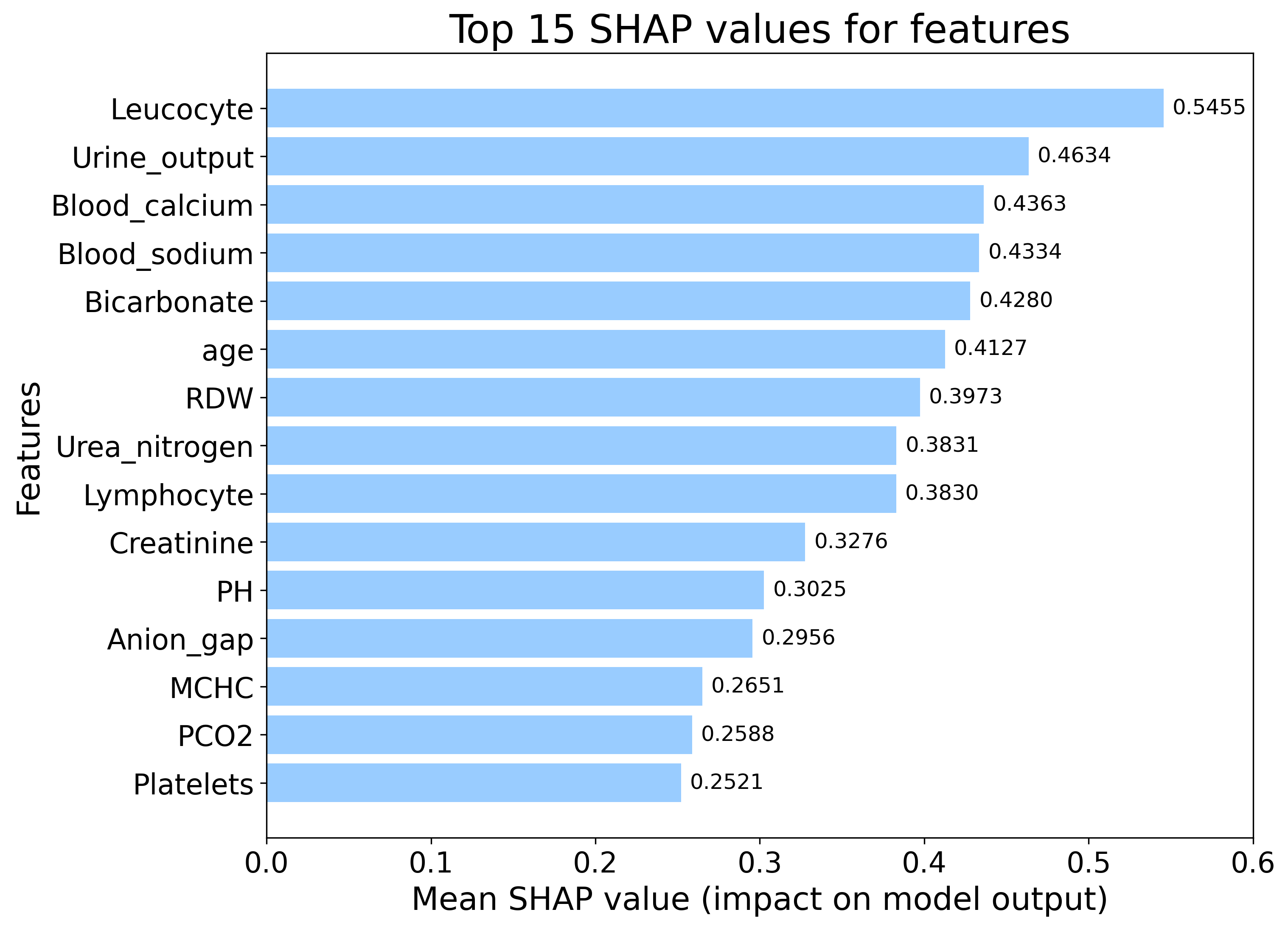}
    \caption{Mean SHAP values for the top 15 features, highlighting 'Leucocyte' as having the highest impact on model predictions.}
    \label{fig:9}
\end{figure*}

\autoref{fig:9} provides a SHAP bar plot showing the mean impact of each feature on the model's predictions. This visualization highlights 'Leucocyte' as having the highest mean SHAP value, reinforcing its critical role in determining patient outcomes. The consistent high SHAP values across different plots indicate the robustness of these features in predicting ICU mortality. These insights are crucial for understanding and interpreting the model's predictions, aiding in the identification of high-risk patients.

\section{DISCUSSION}

\subsection{\textit{Existing model compilation summary}}

Many studies have attempted to predict in-hospital mortality among ICU patients with heart failure using the MIMIC-III database, often facing limitations due to feature selection and imbalanced datasets. For example, Chiu et al. used an ensemble algorithm but struggled with feature selection and data imbalance \cite{ref37}. Our previous study addressed these challenges using advanced imputation methods and targeted feature selection, achieving an LASSO Logistic Regression model with a test AUC-ROC of 0.8766 (95\% CI 0.8065 - 0.9429) and accuracy of 0.729.

In this study, we improved upon our previous work by adding critical features to enhance model performance. Feature extraction was informed by a combination of literature review, clinical expert opinion, and techniques such as VIP and XGBoost, rather than relying on a limited set of features. This approach significantly enhanced predictive performance. Our XGBoost model, incorporating these additional features, achieved an AUC-ROC of 0.9228, marking a significant improvement. This surpasses previous reports, such as that by Li et al., which achieved a best AUC-ROC of 0.824 using XGBoost \cite{ref2}. SHAP analysis further provided clarity on feature importance, enhancing the model's interpretability and its application to clinical decision-making.
 
\subsection{\textit{Study limitations}}

In our model development, we used training and validation datasets from the MIMIC-III database to construct the model, with the test dataset used for performance evaluation. While MIMIC-III offers a large dataset, its data is limited to ICU patients from 2001 to 2012, which may not fully represent current clinical practices. Future work should involve validating the model on independent datasets from various healthcare systems to enhance robustness and generalizability. Additionally, using newer datasets could improve predictive capabilities by aligning with contemporary healthcare trends.

To further enhance model accuracy, integrating other data types such as medical imaging and patient descriptions through Natural Language Processing (NLP) could be highly beneficial. Building NLP models to analyze textual data from patient notes could provide deeper insights into clinical conditions, leading to improved outcome predictions. Furthermore, the development of Neural Network (NN) and deep learning models presents a promising avenue to increase predictive effectiveness, especially when handling large and diverse datasets.

\section{CONCLUSION}

This research developed a machine learning model to predict the mortality of ICU patients with heart failure using data from the MIMIC-III database. We compared several baseline models, including Random Forest, SVM, and KNN, and found that XGBoost outperformed all others, achieving the highest AUC-ROC and the narrowest 95\% confidence intervals, indicating its superior predictive accuracy.

Our approach involved a comprehensive feature selection process, ultimately narrowing the dataset to 46 key features using methods such as VIP, clinical expert opinions, and an ablation study. These steps ensured the selection of the most relevant variables, enhancing model performance. Hyperparameter tuning via Grid-Search optimization further refined the model's capabilities. SHAP analysis validated the clinical significance of important features like leucocyte count and RDW, confirming the robustness and interpretability of the model.

This predictive framework provides essential support for medical professionals by identifying high-risk ICU heart failure patients, facilitating timely and targeted interventions. Such capabilities are crucial in critical care, where accurate and prompt predictions can significantly improve patient outcomes and optimize resource use.
\section*{Acknowledgment}

The authors extend their gratitude to the creators of MIMIC-III for furnishing a thorough and inclusive public electronic health record (EHR) dataset.


\begin{thebibliography}{99}

\bibitem{ref1} S. Emmons-Bell, C. Johnson, and G. Roth, “Prevalence, incidence, and survival of heart failure: a systematic review,” Heart, vol. 108, no. 17, pp. heartjnl-2021-320131, 2022. doi: \href{https://doi.org/10.1136/heartjnl-2021-320131}{10.1136/heartjnl-2021-320131}.

\bibitem{ref2} J. Li, S. Liu, Y. Hu, L. Zhu, Y. Mao, and J. Liu, “Predicting Mortality in Intensive Care Unit Patients With Heart Failure Using an Interpretable Machine Learning Model: Retrospective Cohort Study,” Journal of Medical Internet Research, vol. 24, no. 8, p. e38082, Aug. 2022. doi: \href{https://doi.org/10.2196/38082}{10.2196/38082}.

\bibitem{ref3} B. Bozkurt, T. Ahmad, K. M. Alexander, et al., “Heart Failure Epidemiology and Outcomes Statistics: A Report of the Heart Failure Society of America,” Journal of Cardiac Failure, vol. 29, no. 10, pp. 1412-1451, 2023. doi: \href{https://doi.org/10.1016/j.cardfail.2023.07.006}{10.1016/j.cardfail.2023.07.006}.

\bibitem{ref4} Y. D. Dlugacz, L. Stier, D. Lustbader, M. C. Jacobs, E. Hussain, and A. Greenwood, “Expanding a Performance Improvement Initiative in Critical Care from Hospital to System,” The Joint Commission Journal on Quality Improvement, vol. 28, no. 8, pp. 419–434, Aug. 2002. doi: \href{https://doi.org/10.1016/S1070-3241(02)28042-6}{10.1016/S1070-3241(02)28042-6}.


\bibitem{ref5} A. M. Rahmani, E. Yousefpoor, M. S. Yousefpoor, Z. Mehmood, A. Haider, M. Hosseinzadeh, et al., “Machine Learning (ML) in Medicine: Review, Applications, and Challenges,” Mathematics, vol. 9, no. 22, p. 2970, Nov. 2021. doi: \href{https://doi.org/10.3390/math9222970}{10.3390/math9222970}.

\bibitem{ref6} J. A. M. Sidey-Gibbons and C. J. Sidey-Gibbons, “Machine learning in medicine: a practical introduction,” BMC Medical Research Methodology, vol. 19, no. 1, p. 64, 2019. doi: \href{https://doi.org/10.1186/s12874-019-0681-4}{10.1186/s12874-019-0681-4}.

\bibitem{ref7} American Heart Association, “Heart failure [Internet],” www.heart.org. Available from: \href{https://www.heart.org/en/health-topics/heart-failure}{https://www.heart.org/en/health-topics/heart-failure}.

\bibitem{ref8} R. Miotto, L. Li, B. A. Kidd, and J. T. Dudley, “Deep patient: an unsupervised representation to predict the future of patients from the electronic health records,” Sci Rep, vol. 6, no. 1, p. 26094, 2016. doi: \href{https://doi.org/10.1038/srep26094}{10.1038/srep26094}.

\bibitem{ref9} Y. W. Lin, Y. Zhou, F. A. O. Faghri, M. J. Shaw, and R. H. Campbell, “Analysis and prediction of unplanned intensive care unit readmission using recurrent neural networks with long short-term memory,” PLoS ONE, vol. 14, no. 7, p. e0218942, 2019. doi: \href{https://doi.org/10.1371/journal.pone.0218942}{10.1371/journal.pone.0218942}.

\bibitem{ref10} A. Brnabic and L. M. Hess, “Systematic literature review of machine learning methods used in the analysis of real-world data for patient-provider decision making,” BMC Medical Informatics and Decision Making, vol. 21, no. 1, Feb. 2021. doi: \href{https://doi.org/10.1186/s12911-021-01456-1}{10.1186/s12911-021-01456-1}.

\bibitem{ref11} F. Li, H. Xin, J. Zhang, M. Fu, J. Zhou, and Z. Lian, “Prediction model of in-hospital mortality in intensive care unit patients with heart failure: machine learning-based, retrospective analysis of the MIMIC-III database,” BMJ Open, vol. 11, no. 7, p. e044779, Jul. 2021. doi: \href{https://doi.org/10.1136/bmjopen-2020-044779}{10.1136/bmjopen-2020-044779}.

\bibitem{ref12} M. Kruse, B. Stein, C. Thomas, and S. Kachnowski, “The impact of electronic health records on healthcare quality: a systematic review and meta-analysis,” European Journal of Public Health, vol. 29, no. 5, pp. 1-8, 2019. doi: \href{https://doi.org/10.1093/eurpub/cku015}{10.1093/eurpub/cku015}.

\bibitem{ref13} N. Ashrafi, et al., “Process Mining/Deep Learning Model to Predict Mortality in Coronary Artery Disease Patients,” medRxiv, 2024. doi: \href{https://doi.org/10.1101/2024.06.000000}{10.1101/2024.06.000000}.

\bibitem{ref17} H. Wang, S. W. Park, and J. K. Kim, “Unlocking the power of EHRs: Machine learning for clinical decision support,” Journal of Electrical Systems and Information Technology, vol. 10, 2023, doi: \href{https://doi.org/10.1186/s12938-023-01067-3}{10.1186/s12938-023-01067-3}.


\bibitem{ref14} N. Ashrafi, et al., “Effect of a process mining based pre-processing step in prediction of the critical health outcomes,” arXiv preprint arXiv:2407.02821, 2024. doi: \href{https://doi.org/10.48550/arXiv.2407.02821}{10.48550/arXiv.2407.02821}.

\bibitem{ref18} H. Wang, S. W. Park, and J. K. Kim, “Healthcare predictive analytics using machine learning and deep learning techniques: a survey,” Journal of Electrical Systems and Information Technology, vol. 10, 2023. doi: \href{https://doi.org/10.1186/s12938-023-01067-3}{10.1186/s12938-023-01067-3}.


\bibitem{ref16} A. Brnabic and L. M. Hess, “Systematic literature review of machine learning methods used in the analysis of real-world data for patient-provider decision making,” BMC Medical Informatics and Decision Making, vol. 21, no. 1, Feb. 2021. doi: \href{https://doi.org/10.1186/s12911-021-01456-1}{10.1186/s12911-021-01456-1}.

\bibitem{ref19} J. A. C. Lima, R. N. Rajagopalan, and A. Gupta, “Artificial intelligence and machine learning for cardiovascular disease: Potential applications and perspectives,” Journal of Cardiovascular Computed Tomography, vol. 17, no. 2, 2023. doi: \href{https://doi.org/10.1016/j.jcct.2022.11.006}{10.1016/j.jcct.2022.11.006}.



\bibitem{ref20} C. C. Chiu, C. M. Wu, T. N. Chien, L. J. Kao, C. Li, and H. L. Jiang, “Applying an Improved Stacking Ensemble Model to Predict the Mortality of ICU Patients with Heart Failure,” Journal of Clinical Medicine, vol. 11, no. 21, p. 6460, Jan. 2022. doi: \href{https://doi.org/10.3390/jcm11216460}{10.3390/jcm11216460}.


\bibitem{ref21} Z. Chen, T. Li, S. Guo, D. Zeng, and K. Wang, “Machine learning-based in-hospital mortality risk prediction tool for intensive care unit patients with heart failure,” Frontiers in Cardiovascular Medicine, vol. 10, 2023. doi: \href{https://doi.org/10.3389/fcvm.2023.1119699}{10.3389/fcvm.2023.1119699}.

\bibitem{ref22} Z. Yu, N. Ashrafi, H. Li, K. Alaei, and M. Pishgar, “Prediction of 30-day mortality for ICU patients with Sepsis-3,” BMC Medical Informatics and Decision Making, vol. 24, no. 1, p. 223, 2024. doi: \href{https://doi.org/10.1186/s12911-024-02332-2}{10.1186/s12911-024-02332-2}.

\bibitem{ref23} A. Johnson, T. Pollard, and R. Mark, “MIMIC-III Clinical Database [Internet],” Physionet.org, 2016. Available from: \href{https://physionet.org/content/mimiciii/1.4/}{https://physionet.org/content/mimiciii/1.4/}.

\bibitem{ref28} H. Yun, J. Choi, and J. H. Park, “Prediction of critical care outcome for adult patients presenting to emergency department using initial triage information: an XGBoost algorithm analysis,” JMIR Medical Informatics, vol. 9, no. 9, p. e30770, 2021. doi: \href{https://doi.org/10.2196/30770}{10.2196/30770}.

\bibitem{ref29} N. Ashrafi, A. Abdollahi, and M. Pishgar, “Enhanced Prediction of Ventilator-Associated Pneumonia in Patients with Traumatic Brain Injury Using Advanced Machine Learning Techniques,” arXiv preprint arXiv:2408.01144, 2024. doi: \href{https://doi.org/10.48550/arXiv.2408.01144}{10.48550/arXiv.2408.01144}.

\bibitem{ref25} A. Amritphale, R. Chatterjee, S. Chatterjee, N. Amritphale, A. Rahnavard, G. M. Awan, and G. C. Fonarow, “Predictors of 30-day unplanned readmission after carotid artery stenting using artificial intelligence,” Advances in Therapy, vol. 38, no. 6, pp. 2954–2972, 2021. doi: \href{https://doi.org/10.1007/s12325-021-01701-7}{10.1007/s12325-021-01701-7}.


\bibitem{ref27} M. Pishgar, et al., “Prediction of unplanned 30-day readmission for ICU patients with heart failure,” BMC Medical Informatics and Decision Making, vol. 22, no. 1, p. 117, 2022. doi: \href{https://doi.org/10.1186/s12911-022-01800-3}{10.1186/s12911-022-01800-3}.


\bibitem{ref26} G. S. Collins, D. G. Altman, J. B. Reitsma, K. G. Moons, and the TRIPOD Group, “TRIPOD 2024 statement: updated guidance for reporting clinical prediction models that use regression or machine learning methods,” BMJ, vol. 385, 2024. doi: \href{https://doi.org/10.1136/bmj.q824}{10.1136/bmj.q824}.


\bibitem{ref24} K. G. Moons, D. G. Altman, J. B. Reitsma, J. P. Ioannidis, P. Macaskill, E. W. Steyerberg, and G. S. Collins, “Transparent reporting of a multivariable prediction model for individual prognosis or diagnosis (TRIPOD): explanation and elaboration,” Annals of Internal Medicine, vol. 162, no. 1, pp. W1–W73, 2015. doi: \href{https://doi.org/10.7326/M14-0698}{10.7326/M14-0698}.

\bibitem{ref15} J. Zhang, H. Li, N. Ashrafi, Z. Yu, G. Placencia, and M. Pishgar, “Prediction of in-hospital mortality for ICU patients with heart failure,” medRxiv, 2024. doi: \href{https://doi.org/10.1101/2024.06.000000}{10.1101/2024.06.000000}.



\bibitem{ref30} P. N. Peterson, J. S. Rumsfeld, L. Liang, et al., “A validated risk score for in-hospital mortality in patients with heart failure from the American Heart Association Get With The Guidelines program,” Circ Cardiovasc Qual Outcomes, vol. 3, pp. 25–32, 2010. doi: \href{https://doi.org/10.1161/CIRCOUTCOMES.109.854877}{10.1161/CIRCOUTCOMES.109.854877}.

\bibitem{ref31} Q. Jia, Y.-R. Wang, P. He, et al., “Prediction model of in-hospital mortality in elderly patients with acute heart failure based on retrospective study,” J Geriatr Cardiol, vol. 14, pp. 669–678, 2017. doi: \href{https://doi.org/10.11909/j.issn.1671-5411.2017.11.002}{10.11909/j.issn.1671-5411.2017.11.002}.


\bibitem{ref32} T. Lagu, P. S. Pekow, M. S. Stefan, et al., “Derivation and validation of an in-hospital mortality prediction model suitable for profiling hospital performance in heart failure,” J Am Heart Assoc, vol. 7, 2018. doi: \href{https://doi.org/10.1161/JAHA.116.005256}{10.1161/JAHA.116.005256}.

\bibitem{ref33} N. Wang, R. Gallagher, D. Sze, et al., “Predictors of frequent readmissions in patients with heart failure,” Heart Lung Circ, vol. 28, pp. 277–283, 2019. doi: \href{https://doi.org/10.1016/j.hlc.2017.10.024}{10.1016/j.hlc.2017.10.024}.

\bibitem{ref34} M. O. Akinwande, H. G. Dikko, and A. Samson, “Variance Inflation Factor: As a Condition for the Inclusion of Suppressor Variable(s) in Regression Analysis,” Open Journal of Statistics, vol. 5, no. 7, pp. 754-767, 2015. doi: \href{https://doi.org/10.4236/ojs.2015.57075}{10.4236/ojs.2015.57075}.

\bibitem{ref35} J. Frost, “Independent Samples T-Test: Definition, Examples, Calculator,” Statistics By Jim, 2024. Available from: \href{https://statisticsbyjim.com/hypothesis-testing/independent-samples-t-test-formula-examples/}{https://statisticsbyjim.com/hypothesis-testing/independent-samples-t-test-formula-examples/}.


\bibitem{ref36} R. I. Hamilton and P. N. Papadopoulos, “Using SHAP values and machine learning to understand trends in the transient stability limit,” IEEE Transactions on Power Systems, vol. 39, no. 1, pp. 1384-1397, 2023. doi: \href{https://doi.org/10.1109/TPWRS.2022.1234567}{10.1109/TPWRS.2022.1234567}.

\bibitem{ref37} C.-C. Chiu, C.-M. Wu, T.-N. Chien, L.-J. Kao, C. Li, and H.-L. Jiang, “Applying an improved stacking ensemble model to predict the mortality of ICU patients with heart failure,” Journal of Clinical Medicine, vol. 11, no. 21, p. 6460, 2022. doi: \href{https://doi.org/10.3390/jcm11216460}{10.3390/jcm11216460}.


\end{thebibliography}
\end{document}